%% file: main.tex
\documentclass[a4paper,twoside]{article}

\input{preamble}

\begin{document}

\title{Fruit-HSNet: A Machine Learning Approach for Hyperspectral Image-Based Fruit Ripeness Prediction}

\input{authors}

\input{sections/00_abstract}

\addpublicationnotice

\onecolumn
\maketitle
\normalsize
\setcounter{footnote}{0}
\vfill

\input{sections/01_introduction}
\input{sections/02_methodology}
\input{sections/03_experimental_evaluation}
\input{sections/04_ablation_study}

\input{sections/05_discussion}
\input{sections/06_conclusion}
\input{sections/07_acknowledgements}

\bibliographystyle{apalike}
{\small
  \bibliography{references}
}

\end{document}

%% file: preamble.tex
\usepackage{epsfig}
\usepackage{calc}
\usepackage{amstext,amssymb,amsmath,amsthm}
\usepackage{multicol}
\usepackage{pslatex}
\usepackage{apalike}
\usepackage{algorithm2e}
\usepackage[bottom]{footmisc}

\usepackage{graphicx}
\usepackage{array}
\usepackage{booktabs}
\usepackage{multirow}
\usepackage{pifont}
\usepackage[font=small]{caption}
\usepackage{subfig}
\usepackage[table]{xcolor}

\definecolor{LightGray}{gray}{0.9}

\usepackage{pgfplots}
\pgfplotsset{compat=1.16}

\usepackage[
  pagebackref=false,
  breaklinks=true,
  colorlinks=true,
  citecolor=blue,
  bookmarks=false
]{hyperref}

\newcommand{\paragraphheading}[1]{%
  \vspace{.05in}\noindent\textbf{#1}%
}

\usepackage{eso-pic}
\usepackage{ragged2e}

\newcommand{\addpublicationnotice}{%
  \AddToShipoutPictureFG*{%
    \AtTextLowerLeft{%
      \raisebox{-15mm}[0pt][0pt]{%
        \parbox[b]{\textwidth}{%
          \fontsize{6.5}{7.5}\selectfont
          \itshape
          \justifying
          \noindent
        This is the authors' manuscript. The Version of Record was published in the
        Proceedings of the 17th International Conference on Agents and Artificial
        Intelligence (ICAART 2025), Vol.~2, pp.~102--111, SciTePress, 2025.
        The Version of Record is available at
        \href{https://doi.org/10.5220/0013110800003890}
        {https://doi.org/10.5220/0013110800003890}
        and is licensed under the Creative Commons
        Attribution-NonCommercial-NoDerivatives 4.0 International license
        (CC BY-NC-ND 4.0).%
          \par
        }%
      }%
    }%
  }%
}

\usepackage{SCITEPRESS}

%% file: authors.tex
\author{\authorname{Ahmed Baha Ben Jmaa\sup{1}\orcidAuthor{0009-0006-0333-2643}, Faten Chaieb\sup{1}\orcidAuthor{0000-0002-2968-2426}, and Anna Fabijańska\sup{2}\orcidAuthor{0000-0002-0249-7247}}
\affiliation{\sup{1}Efrei Research Lab, Paris Panthéon-Assas University, Paris, France}
\affiliation{\sup{2}Institute of Applied Computer Science, Lodz University of Technology, Łódź, Poland}
\email{ahmedbaha.benjmaa@outlook.fr, faten.chakchouk@efrei.fr, anna.fabijanska@p.lodz.pl,}
}

%% file: sections/00_abstract.tex
\keywords{Fruit Ripeness Prediction, Hyperspectral Image, DeepHS Fruit Dataset, Smart Agriculture}
\abstract{
Fruit ripeness prediction (FRP) is a classification-based agricultural computer vision task that has attracted much attention, thanks to its wide-ranging advantages in agriculture field for both pre-harvest and post-harvest management. Accurate and timely FRP can be achieved using machine/deep learning-based hyperspectral image classification techniques. However, challenges including the limited availability of labeled data and the lack of robust methods generalizable to various hyperspectral cameras and fruit types can compromise the effectiveness of hyperspectral image-based FRP.
Addressing these challenges, this paper introduces Fruit-HSNet, a machine learning architecture specifically designed for hyperspectral classification of fruit ripeness. Fruit-HSNet incorporates a spatio-spectral feature extraction module based on Fourier Transform and central pixel spectral signature followed by learnable feature fusion and a classifier optimized for ripeness classification.
The proposed architecture was evaluated using the DeepHS Fruit dataset, the largest publicly available labeled real-world hyperspectral dataset for predicting fruit ripeness, which includes five different types of fruits—avocado, kiwi, mango, kaki, and papaya—captured with three distinct hyperspectral cameras at various stages of ripeness. Experimental results highlight that Fruit-HSNet substantially outperforms existing deep learning methods, from baseline to state-of-the-art models, with improvements of 12\%, achieving a new state-of-the-art overall accuracy~of~70.73\%.
}

%% file: sections/01_introduction.tex
\section{\uppercase{Introduction}}
In the field of smart agriculture, agricultural computer vision is attracting increasing attention for various applications, from irrigation management to automated classification of agricultural products, enabling automated and simplified agricultural tasks~\cite{GHAZAL202464,luo2023survey,LU2020105760}. Fruit Ripeness Prediction (FRP) is an agricultural computer vision task that involves classifying fruits to their degree of ripeness, offering several advantages for both pre-harvest and post-harvest management, including minimizing losses, improving quality, and economizing resources~\cite{Rizzo2023}.

Traditionally, FRP has been performed using methods such as visual observation and chemical analysis of the fruit. However, these techniques are subjective, labor-intensive, and costly, involving a significant margin of error while consuming human and material resources. The emergence of machine/deep learning and imaging technologies, including hyperspectral imaging, has enabled the development of new FRP methods by leveraging the power of learning algorithms to learn hidden patterns. These methods offer advantages over traditional methods, such as the ability to make accurate and timely predictions~\cite{Rizzo2023,RAM2024109037}.

Hyperspectral imaging (HSI), in particular, unlike conventional imaging techniques, offers the advantage of capturing spatial and spectral information across a wide range of the electromagnetic spectrum, providing details not visible to humans. Formally, a hyperspectral image $\mathbf{H} \in \mathbb{R}^{M \times N \times B}$ is defined as a three-dimensional data cube with two spatial dimensions, $M$ and $N$, representing spatial information, and one spectral dimension, $B$, representing spectral information (i.e., wavelength), encapsulating the reflectance properties of the materials present in the image at different wavelengths. The intensity value of each pixel at spatial coordinates $(x, y)$ and wavelength $\lambda$ corresponding to a specific spectral band can be described as $\mathbf{H}(x,y,\lambda) = r_{\lambda}$, where $r_{\lambda}$ denotes the spectral response or reflectance at that wavelength. The entire spectral response $\mathbf{H(x,y,:)} = \mathbf{r}$ for a pixel at the spatial coordinates~$(x, y)$ represents the spectral reflectance curve of the object at that location, encompassing its full spectral profile. These spectral reflectance curves are essential for distinguishing materials based on their unique spectral properties, often related to their chemical composition and structure~\cite{Muhammad2022}.

Hyperspectral image classification (HIC) has been widely studied in the literature, and various methods have been proposed~\cite{Kumar2024,ahmad2024,Muhammad2022}. These range from traditional machine learning methods such as Support Vector Machines~(SVM), k-Nearest Neighbors~(KNN), and dimensionality reduction techniques, to deep learning methods based on convolution and attention techniques. However, the development of these methods is application-specific aware, which limits their generalizability. This makes the adaptation of HIC to new applications a significant challenge. Indeed, while state-of-the-art methods for HIC have shown impressive results in certain applications, they fail to maintain comparable performance across different applications~\cite{frank2023hyperspectral}.

In this context, several works and datasets have been proposed for hyperspectral image-based FRP~\cite{Zhu2017,Barrera2019,VargaMZ21,VargaFZ23,frank2023hyperspectral,Rizzo2023}. Principally, the DeepHS Fruit dataset~\cite{VargaMZ21,VargaFZ23} and the DeepHS-Net family of architectures~\cite{VargaMZ21,VargaFZ23,VargaMBZ23} represent the state-of-the-art. 
The DeepHS Fruit dataset is the largest commonly available real-world hyperspectral dataset for FRP, distinguished by its variety in the number of fruits, types of hyperspectral cameras used, and stages of maturity. The DeepHS-Net family of architectures, a convolution-based deep learning methods, is specifically designed for hyperspectral classification of fruit ripeness. This family includes two principal convolutional neural network (CNN) architectures: \textbf{(1)}~\textit{DeepHS-Net}~\cite{VargaMZ21}, which uses depthwise separable 2D convolutions, and~\textbf{(2)}~\textit{DeepHS-Hybrid-Net}~\cite{VargaFZ23}, which combines 2D and 3D depthwise separable convolutions. Additional variants derived from these two architectures incorporate HyveConv (Hyperspectral Visual Embedding Convolution), a wavelength-aware 2D convolution~\cite{VargaMBZ23}. HyveConv employs a continuous representation of convolution kernels, sampling these kernels based on the wavelengths of the inputs. This design makes the convolution independent of the camera type used and efficiently reduces the number of parameters.

Despite the performance demonstrated by DeepHS-Net architectures in FRP, significant challenges remain unresolved. The lack of robust methods that can generalize across different hyperspectral cameras and fruit types, with the limited size of datasets, compromises the effectiveness of hyperspectral image-based FRP.
In response to these limitations, this paper introduces Fruit-HSNet, an architecture specifically designed for FRP from hyperspectral images. The main objective is to ensure consistent and accurate classification across different hyperspectral cameras, fruit types, and stages of ripeness.

\paragraphheading{Contributions.}
Our key contributions are summarized as follows:
\textbf{(1)}~We propose Fruit-HSNet, a new architecture specifically designed for fruit hyperspectral image classification to identify different stages of fruit ripeness by leveraging spatio-spectral descriptors, which include Fourier Transform-based features, central pixel spectral signatures, and learnable feature fusion.
\textbf{(2)}~We conducted comprehensive evaluations on the DeepHS Fruit dataset, the largest publicly available labeled hyperspectral dataset for fruit maturity prediction, which includes five different types of fruits—avocado, kiwi, mango, kaki, and papaya—captured using three distinct hyperspectral cameras.
\textbf{(3)}~We demonstrate that Fruit-HSNet achieved a new state-of-the-art overall accuracy of 70.73\% on the DeepHS Fruit benchmark dataset, which is a 12\% improvement over the previous best-reported results, specifically in the challenging categories of avocados and kiwis, which are critical due to their ripening processes.

\paragraphheading{Paper Organization.} 
In the following, Section~\ref{sec:methodology} details our methodology, including an exploratory analysis of the DeepHS Fruit dataset and the introduction of the Fruit-HSNet architecture. Subsequent sections evaluate the model's performance, analyze the results, and discuss conclusions along with future research directions.

%% file: sections/02_methodology.tex
\begin{figure*}
    \centering
    \includegraphics[width=\linewidth]{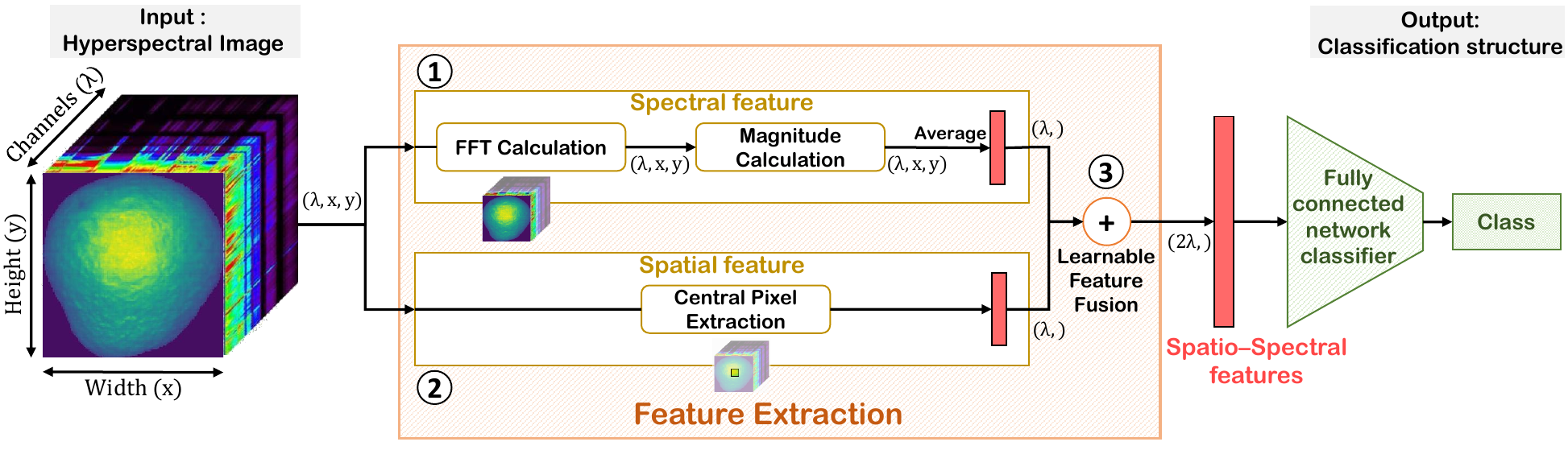}
    \caption{Illustration of Fruit-HSNet architecture.}
\end{figure*}

\section{\uppercase{Methodology}}
\label{sec:methodology}
    \subsection{DeepHS Fruit Dataset}
    \label{sec:deep-hs-fruit-dataset}
    DeepHS Fruit dataset~\cite{VargaMZ21,VargaFZ23} is the largest publicly available real-world hyperspectral dataset labeled for fruit ripeness prediction. This dataset includes hyperspectral images of five different fruit types, captured by three distinct hyperspectral cameras, and categorized according to various ripeness level.
    
    \paragraphheading{Dataset composition.} The DeepHS Fruit dataset consists of 30 configurations, where each configuration, denoted as \textit{$\text{{config}}_i$}, corresponds to the dataset for a specific fruit \textit{$\text{{fruit}}_i$} within the category \textit{$\text{{categorie}}_i$}, captured with the camera \textit{$\text{{camera}}_i$}. The fruits (\textit{$\text{{fruit}}_i$}) included are avocado, kiwi, mango, kaki, and papaya. Fruit ripeness in this dataset is classified into three distinct categories (\textit{$\text{{categorie}}_i$}): ripeness, firmness, and sweetness. The cameras (\textit{$\text{{camera}}_i$}) used are the Specim FX 10, Corning microHSI 410, and Innospec Redeye, with the following exceptions:
    \begin{itemize}
        \item No sweetness category for avocados.
        \item No records captured with Innospec Redeye camera for mango, kaki, and papaya.
        \item No records captured with Corning microHSI camera for kiwi.
    \end{itemize}
    
    \paragraphheading{Class labels.} For each \textit{$\text{{config}}_i$}, three classes are per category, defined as follows:
    \begin{itemize}
        \item \textbf{Ripeness:} unripe, ripe, overripe.
        \item \textbf{Firmness:} too firm, perfect, too soft.
        \item \textbf{Sweetness:} not sweet, sweet, overly sweet.
    \end{itemize}
    
    \paragraphheading{Data collection.}
    In addition to the hyperspectral image label, metadata are available including type of fruit, orientation (front or back), capturing camera, and wavelengths of the recorded spectra. The cameras vary in their spectral band capture:
    \begin{itemize}
        \item \textit{Camera 1: Specim FX 10} captures 224 spectral bands with a wavelength range of 400 to 1000 nm.
        \item \textit{Camera 2: Corning microHSI 410} captures 249 spectral bands with a wavelength range of 920 to 1730 nm.
        \item \textit{Camera 3: Innospec Redeye} captures 252 spectral bands, also spanning 920 to 1730 nm.
    \end{itemize}
    
    \paragraphheading{Dataset distribution.}
    DeepHS Fruit dataset comprises a total of 2706 labeled hyperspectral images, distributed among the fruits as follows: 461 images for Avocado, 568 images for Kiwi, 336 images for Mango, 336 images for Kaki, and 252 images for Papaya.

    \subsection{Fruit-HSNet: Proposed Method}
    This section introduces the architecture of Fruit-HSNet and its working principle for fruit ripeness classification. Let $\mathbf{H} \in \mathbb{R}^{M \times N \times B}$ represent a hyperspectral image, the input of the Fruit-HSNet architecture, where $B$ is the number of spectral channels, and $M$ and $N$ are the height and width of the image, respectively. Fruit-HSNet extracts both spectral and spatial features from $\mathbf{H}$, which are informative and discriminative, via a feature extraction module and then classifies them via a classification module based on a fully connected neural network.
    
    Feature extraction is performed through a dual-branch approach: \textbf{(1)} Spectral Feature Extraction, \textbf{(2)} Spatial Feature Extraction, followed by \textbf{(3)} Learnable Feature Fusion.
    
    \paragraphheading{(1) Spectral Feature Extraction Module.} 
    In this branch, a Fourier Transform (FT) is applied to $\mathbf{H}$ to transform the spatial information $\mathbf{H}(:,:,\lambda)$ of each spectral channel $\lambda$ into the frequency domain: 
    \begin{equation} 
    \mathbf{F}_\lambda = \text{\textbf{FT}}\left(\mathbf{H}\left(:,:,\lambda\right)\right), \quad 1 \leq \lambda \leq B 
    \end{equation} 
    After applying FT across the spatial dimensions, the magnitudes of $\mathbf{F}_\lambda$ are computed and averaged over each channel $\lambda$ to form $\mathbf{m}$. This frequency transformation makes the periodic patterns of textures and structural changes in the fruit skin more discernible. Indeed, changes in the fruit during ripening affect its spectral signature. By transforming features into the frequency domain, Fruit-HSNet can effectively capture patterns associated with various stages of fruit ripeness that might be less evident in the spatial domain.
    
    \paragraphheading{(2) Spatial Feature Extraction Module.} 
    In this branch, the central pixel spectral signature, $\mathbf{s} = [s_\lambda, 1 \leq \lambda \leq B]$, is extracted, where 
    \begin{equation} 
    s_\lambda = \mathbf{H}\left(\frac{M}{2}, \frac{N}{2}, \lambda\right), \quad 1 \leq \lambda \leq B 
    \end{equation} 
    This allows focusing on potentially the most chemically informative region of the fruit, which is generally indicative of its overall ripeness.
    
    \paragraphheading{(3) Learnable Feature Fusion.} 
    Features output from the two branches are adaptively weighted by learnable parameters $\mathbf{w_{1}}, \mathbf{w_{2}} \in \mathbb{R}^B$. The weighted features are then concatenated to form a feature vector $\mathbf{f}$ that combines both spectral and spatial information. 
    \begin{equation} 
    \mathbf{f} = \mathbf{w_{1}} \mathbf{m} + \mathbf{w_{2}} \mathbf{s} 
    \end{equation} 
    The introduction of learnable weights for each feature type allows Fruit-HSNet to adaptively prioritize which type of feature (spectral or spatial) is more informative based on their relevance to fruit ripeness. The adaptability provided by these weights enables the model to be applied effectively across different types of fruits, varying stages of ripeness, and different camera types, where the importance of spectral versus spatial information may differ.

%% file: sections/03_experimental_evaluation.tex
\section{\uppercase{Experiments and Results}}
    \subsection{Experimental Setup}
    Experiments were conducted using 30 datasets from the DeepHS Fruit dataset~\cite{VargaMZ21,VargaFZ23} as described in Section~\ref{sec:deep-hs-fruit-dataset}, adhering to the standard data splitting and preprocessing procedures outlined in~\cite{frank2023hyperspectral} to ensure result comparability. 
    
    Fruit-HSNet was trained on each dataset with the cross-entropy loss function, using a batch size of 16. Optimization involved the use of the Adam algorithm, initiated with a learning rate of 0.001 alongside a weight decay factor of $1 \times 10^{-4}$. Additionally, a learning rate scheduler reduced the rate by a factor of 0.7 every 10 epochs to fine-tune the training process. Furthermore, a variable number of epochs is determined experimentally for each configuration to avoid overfitting.
    
    Fruit-HSNet's fusion feature module uses learnable weights, initialized using a normal distribution. In the classification module, the fully connected neural network comprises two linear layers with dimensions $[2 \lambda, 512, 256]$, where $\lambda \in \{224, 249, 252\}$. Following these layers, batch normalization is applied, along with ReLU (Rectified Linear Unit) activation functions and dropout layers, implemented with a dropout rate of 0.4.
    
    \subsection{Comparison with State-of-the-Art}
    This section presents a comparative performance evaluation of Fruit-HSNet for hyperspectral classification of fruit ripeness. The evaluation covers specific aspects of performance, starting with a general evaluation across all fruits, hyperspectral cameras, and stages of ripeness in Section~\ref{sec:global-performance}. It is followed by detailed evaluations focusing on fruit-specific and camera-specific variations in Sections~\ref{sec:fruit-specific-evaluation} and~\ref{sec:camera-specific-evaluation}, respectively. The section concludes with an in-depth performance analysis for two critical case studies: avocados and kiwis.
     \begin{table}
        \centering
        \caption{Comparative performance of Fruit-HSNet and state-of-the-art methods across all fruits, hyperspectral cameras, and stages of ripeness on DeepHS Fruit dataset~\cite{VargaMZ21,VargaFZ23}. \textit{$^\ast$Results were obtained from~\cite{frank2023hyperspectral}.}}
        \label{tab:overall-performance}
        \resizebox{\columnwidth}{!}{
        \begin{tabular}{llr}
            \toprule
             \multicolumn{2}{l}{\textbf{Method}} & \textbf{Accuracy} \\ 
            \midrule
            \multirow{16}{*}{\rotatebox{90}{\textbf{Baseline Methods}$^*$}}&\multicolumn{2}{c}{\textit{Convolution-based Methods}}\\
            \cmidrule(lr){2-3}
             &2D CNN (spatial)~\cite{PaolettiHPP19} &44.85 \%\\
             &ResNet-152~\cite{HeZRS16} &47.00 \%\\
             &HybridSN~\cite{RoyKDC20} &48.74 \%\\ 
             &ResNet-18~\cite{HeZRS16} &49.05 \%\\
             &SpectralNET~\cite{Chakraborty2021SpectralNETES} &49.25 \%\\
             &2D CNN (spectral)~\cite{frank2023hyperspectral} &49.27 \%\\
             &1D CNN~\cite{PaolettiHPP19} &51.30 \%\\ 
             &Gabor CNN~\cite{GhamisiMLSTM18} &52.57 \%\\
             &EMP CNN~\cite{GhamisiMLSTM18} &52.76 \%\\ 
             &2D CNN~\cite{PaolettiHPP19} &54.42 \%\\
             &3D CNN~\cite{PaolettiHPP19} &56.06 \%\\ 
             \cmidrule(lr){2-3}
             &\multicolumn{2}{c}{\textit{Attention/Transformer-based Methods}}\\
             \cmidrule(lr){2-3}
             &SpectralFormer~\cite{HongHYGZPC22} &41.71 \%\\
             &Attention-based CNN~\cite{LorenzoTMN20} &44.88 \%\\
             &HiT~\cite{YangCLZ22}  &48.16 \%\\
            \midrule
            \multirow{5}{*}{\rotatebox{90}{\textbf{SOTA Methods}}}&\multicolumn{2}{c}{\textit{DeepHS-Net Family}}\\
            \cmidrule(lr){2-3}
             &DeepHS-Hybrid-Net~\cite{VargaFZ23} &55.01 \%\\
             &DeepHS-Net+HyveConv~\cite{VargaMBZ23} &57.57 \%\\ 
             &DeepHS-Net~\cite{VargaMZ21} &\underline{58.28 \%}\\ 
             \cmidrule(lr){2-3}
             & \cellcolor{LightGray}\textbf{Fruit-HSNet (Ours)} &\cellcolor{LightGray}\textbf{70.73} \%\\
            \bottomrule
        \end{tabular}
        }
    \end{table}
    
    \subsubsection{Global Performance Evaluation}
    \label{sec:global-performance}
    In this part of the evaluation, we compared Fruit-HSNet across all fruits, hyperspectral cameras, and stages of ripeness against baseline and existing state-of-the-art methods. Baseline methods encompass deep learning models for hyperspectral image classification, ranging from convolutional to attention mechanisms, including 2D and 3D CNN variants, and adapted transformer architectures. The state-of-the-art models are represented by the DeepHS-Net family of convolution-based methods. 
    
    In Table~\ref{tab:overall-performance}, we report the overall classification accuracy for each method. This accuracy metric is calculated as the average across 30 diverse datasets, each representing a unique combination of fruit type, camera type, and ripeness category, thus providing a robust measure of model generalizability and effectiveness.
    Subsequently, the DeepHS-Net method is used for detailed benchmarking as it is the best-performing competitor.
    Furthermore, to provide a detailed analysis of performance, various classification metrics, including accuracy, precision, recall, F1-score, and Cohen's Kappa ($\kappa$), were reported in Figure~\ref{fig:overall-performance} to compare Fruit-HSNet against the DeepHS-Net method.
    \input{figures/plots/overall_performance}

    \subsubsection{Fruit-Specific Performance Evaluation}
    \label{sec:fruit-specific-evaluation}
    \input{figures/plots/stratified_performance}
    In this section, performance was analyzed for each fruit type. An overview of the classification performance was presented in Table~\ref{tab:fruit-specific-performance} and Figure~\ref{fig:performance-by-fruit}, followed by a detailed analysis for each fruit in each category~(ripeness, fruitiness, sweetness) in Table~\ref{tab:fruit-specific-details}.
    \begin{table}
        \centering
        \caption{Fruit-specific performance comparison of Fruit-HSNet.}
        \label{tab:fruit-specific-performance}
        \resizebox{\columnwidth}{!}{
        \begin{tabular}{llr>{\columncolor{LightGray}}r}
            \toprule
             \multicolumn{2}{c}{}& DeepHS-Net&\textbf{Fruit-HSNet (Ours)} \\ 
             \midrule
            \multirow{4}{*}{\textbf{Avocado}}& Overall Accuracy  & 77.62\%&85.19\% (\textcolor{blue}{$\uparrow$ 7.57\%})\\ 
             & Average Accuracy   & 77.62\%&82.91\% (\textcolor{blue}{$\uparrow$ 5.29\%})\\ 
             & F1-score   & 76.22\%&84.26\% (\textcolor{blue}{$\uparrow$ 8.04\%})\\
             & $\kappa$   & 66.03\%&76.52\% (\textcolor{blue}{$\uparrow$ 10.49\%})\\ 
             \midrule
            \multirow{4}{*}{\textbf{Kiwi}}& Overall Accuracy  & 60.11\%&71.23\% (\textcolor{blue}{$\uparrow$ 11.12\%})\\ 
             & Average Accuracy   & 60.11\%&63.69\% (\textcolor{blue}{$\uparrow$ 3.58\%})\\ 
             & F1-score   & 58.02\%&68.21\% (\textcolor{blue}{$\uparrow$ 10.19\%})\\
             & $\kappa$   & 36.36\%&52.73\% (\textcolor{blue}{$\uparrow$ 16.37\%})\\ 
             \midrule
            \multirow{4}{*}{\textbf{Mango}}& Overall Accuracy  & 42.59\%&65.28\% (\textcolor{blue}{$\uparrow$ 22.69\%})\\ 
             & Average Accuracy   & 42.59\%&60.83\% (\textcolor{blue}{$\uparrow$ 18.24\%})\\ 
             & F1-score   & 34.80\%&63.73\% (\textcolor{blue}{$\uparrow$ 28.93\%})\\
             & $\kappa$   & 3.51\%&44.86\% (\textcolor{blue}{$\uparrow$ 41.35\%})\\
             \midrule
            \multirow{4}{*}{\textbf{Kaki}}& Overall Accuracy  & 51.85\%&59.72\% (\textcolor{blue}{$\uparrow$ 7.87\%})\\ 
             & Average Accuracy   &44.87\%&52.13\% (\textcolor{blue}{$\uparrow$ 7.26\%})\\ 
             & F1-score   & 42.03\%&51.08\% (\textcolor{blue}{$\uparrow$ 9.05\%})\\
             & $\kappa$   & 21.37\%&33.83\% (\textcolor{blue}{$\uparrow$ 12.46\%})\\ 
             \midrule
            \multirow{4}{*}{\textbf{Papaya}}& Overall Accuracy  & 62.96\%&72.23\% (\textcolor{blue}{$\uparrow$ 9.27\%})\\ 
             & Average Accuracy   & 55.64\%&66.30\% (\textcolor{blue}{$\uparrow$ 10.66\%})\\ 
             & F1-score   & 55.90\%&70.17\% (\textcolor{blue}{$\uparrow$ 14.27\%})\\
             & $\kappa$   & 35.51\%&52.69\% (\textcolor{blue}{$\uparrow$ 17.18\%})\\ 
            \bottomrule
        \end{tabular}
        }
    \end{table}

    \begin{table}
        \centering
        \caption{Detailed fruit-specific performance comparison of Fruit-HSNet across ripeness, firmness, and sweetness}
        \label{tab:fruit-specific-details}
        \resizebox{\columnwidth}{!}{
        \begin{tabular}{llr>{\columncolor{LightGray}}r}
            \toprule
             \multicolumn{2}{c}{}& DeepHS-Net&\textbf{Fruit-HSNet (Ours)} \\ 
             \midrule
            \multirow{5}{*}{\textbf{Ripeness}}& Avocado  & 77.16\%&87.04\% (\textcolor{blue}{$\uparrow$ 9.88\%})\\ 
             & Kiwi   & 57.87\%&79.86\% (\textcolor{blue}{$\uparrow$ 21.99\%})\\ 
             & Mango   & 41.67\%&54.17\% (\textcolor{blue}{$\uparrow$ 12.50\%})\\ 
             & Kaki   & 45.84\%&50.00\% (\textcolor{blue}{$\uparrow$ 4.17\%})\\ 
             & Papaya   & 51.85\%&77.78\% (\textcolor{blue}{$\uparrow$ 25.93\%})\\ 
            \midrule
            \multirow{5}{*}{\textbf{Firmness}}& Avocado  & 78.09\%&83.33\% (\textcolor{blue}{$\uparrow$ 5.25\%})\\ 
             & Kiwi   &63.61\%&72.47\% (\textcolor{blue}{$\uparrow$ 8.86\%})\\ 
             & Mango   & 43.06\%&70.83\% (\textcolor{blue}{$\uparrow$ 27.77\%})\\ 
             & Kaki   & 63.89\%&66.67\% (\textcolor{blue}{$\uparrow$ 2.77\%})\\ 
             & Papaya   & 70.37\%&77.78\% (\textcolor{blue}{$\uparrow$ 7.41\%})\\ 
            \midrule
            \multirow{4}{*}{\textbf{Sweetness}}& Kiwi   & 58.86\%&61.35\% (\textcolor{blue}{$\uparrow$ 2.49\%})\\ 
             & Mango   & 43.06\%&70.84\% (\textcolor{blue}{$\uparrow$ 27.78\%})\\ 
             & Kaki   & 45.83\%&62.50\% (\textcolor{blue}{$\uparrow$ 16.67\%})\\ 
             & Papaya   & 66.67\%&61.12\% (\textcolor{red}{$\downarrow$ 5.55\%})\\ 
            \bottomrule
        \end{tabular}
        }
    \end{table}
    
    \subsubsection{Camera-Specific Performance Evaluation}
    \label{sec:camera-specific-evaluation}
    To assess the robustness of Fruit-HSNet across various hyperspectral cameras, we evaluated its performance by presenting the behavior of classification performance in a global manner in Table~\ref{tab:camera-specific-performance} and Figure~\ref{fig:performance-by-camera}, and in detailed form for each fruit ripeness category in Table~\ref{tab:camera-specific-details}.
    \begin{table}
        \centering
        \caption{Camera-specific performance comparison of Fruit-HSNet.}
        \label{tab:camera-specific-performance}
        \resizebox{\columnwidth}{!}{
        \begin{tabular}{llr>{\columncolor{LightGray}}r}
            \toprule
             \multicolumn{2}{c}{}& DeepHS-Net&\textbf{Fruit-HSNet (Ours)} \\ 
             \midrule
            \multirow{4}{*}{\textbf{Camera 1}}& Overall Accuracy  & 58.83\%&69.81\% (\textcolor{blue}{$\uparrow$ 10.98\%})\\ 
             & Average Accuracy   & 52.79\%&63.98\% (\textcolor{blue}{$\uparrow$ 11.19\%})\\ 
             & F1-score   & 53.12\%&67.12\% (\textcolor{blue}{$\uparrow$ 14\%})\\
             & $\kappa$   &31.42\%&51.63\% (\textcolor{blue}{$\uparrow$ 20.21\%})\\ 
             \midrule
            \multirow{4}{*}{\textbf{Camera 2}}& Overall Accuracy  & 61.20\%&71.72\% (\textcolor{blue}{$\uparrow$ 10.52\%})\\ 
             & Average Accuracy   & 55.01\%&65.50\% (\textcolor{blue}{$\uparrow$ 10.49\%})\\ 
             & F1-score   & 54.92\%&67.67\% (\textcolor{blue}{$\uparrow$ 12.75\%})\\
             & $\kappa$   & 35.19\%&52.09\% (\textcolor{blue}{$\uparrow$ 16.9\%})\\ 
             \midrule
            \multirow{4}{*}{\textbf{Camera 3}}& Overall Accuracy  & 54.81\%&71.11\% (\textcolor{blue}{$\uparrow$ 16.3\%})\\ 
             & Average Accuracy   & 52.03\%&67.78\% (\textcolor{blue}{$\uparrow$ 15.75\%})\\ 
             & F1-score   & 50.82\%&68.15\% (\textcolor{blue}{$\uparrow$ 17.33\%})\\
             & $\kappa$   &29.90\%&53.60\% (\textcolor{blue}{$\uparrow$ 23.70\%})\\
            \bottomrule
        \end{tabular}
        }
    \end{table}

     \begin{table}
        \centering
        \caption{Detailed camera-specific performance comparison of Fruit-HSNet across ripeness, firmness, and sweetness.}
        \label{tab:camera-specific-details}
        \resizebox{\columnwidth}{!}{
        \begin{tabular}{llr>{\columncolor{LightGray}}r}
            \toprule
             \multicolumn{2}{c}{}& DeepHS-Net&\textbf{Fruit-HSNet (Ours)} \\ 
             \midrule
            \multirow{3}{*}{\textbf{Ripeness}}& Camera 1  & 53.70\%&61.94\% (\textcolor{blue}{$\uparrow$ 8.24\%})\\ 
             & Camera 2   & 61.57\%&74.30\% (\textcolor{blue}{$\uparrow$ 12.73\%})\\ 
             & Camera 3   & 55.55\%&88.89\% (\textcolor{blue}{$\uparrow$ 33.33\%})\\ 
            \midrule
            \multirow{3}{*}{\textbf{Firmness}}& Camera 1  & 64.70\%&76.76\% (\textcolor{blue}{$\uparrow$ 12.06\%})\\ 
             & Camera 2   & 69.44\%&77.08\% (\textcolor{blue}{$\uparrow$ 7.64\%})\\ 
             & Camera 3   & 57.40\%&66.67\% (\textcolor{blue}{$\uparrow$ 9.26\%})\\ 
            \midrule
            \multirow{3}{*}{\textbf{Sweetness}}& Camera 1  & 57.90\%&70.95\% (\textcolor{blue}{$\uparrow$ 13.05\%})\\ 
             & Camera 2   & 49.69\%&61.11\% (\textcolor{blue}{$\uparrow$ 11.42\%})\\ 
             & Camera 3   & 48.15\%&44.44\% (\textcolor{red}{$\downarrow$ 3.71\%})\\ 
            \bottomrule
        \end{tabular}
        }
    \end{table}
     
    \subsubsection{Detailed Performance Evaluation for Avocado and Kiwi}
    As avocados and kiwis have a delicate ripeness cycle, this section details the performance of these fruits by camera type and by ripeness category. A detailed analysis of the two categories, ripeness and firmness, is presented in Tables~\ref{tab:avocado-kiwi-ripeness} and~\ref{tab:avocado-kiwi-firmness}, including standard classification metrics (accuracy, F1-score, and $\kappa$).
     \begin{table}
        \centering
        \caption{Detailed performance evaluation for avocado and kiwi (Ripeness categorie)}
        \label{tab:avocado-kiwi-ripeness}
        \resizebox{\columnwidth}{!}{
        \begin{tabular}{lllr>{\columncolor{LightGray}}r}
            \toprule
             &&& DeepHS-Net&\textbf{Fruit-HSNet (Ours)} \\ 
             \midrule
             \multirow{9}{*}{\textbf{Avocado}}  &\multirow{3}{*}{Camera 1} &Accuracy&83.33\%&83.33\% (\textcolor{blue}{$\uparrow$ 0\%})\\ 
               & &F1-score & 82.94\%&83.20\% (\textcolor{blue}{$\uparrow$ 0.26\%})\\ 
                & &$\kappa$ & 75.00\%&75.00\% (\textcolor{blue}{$\uparrow$ 0\%})\\ 
            \cmidrule(lr){2-5}
                &\multirow{3}{*}{Camera 2}&Accuracy& 88.89\%&88.89\% (\textcolor{blue}{$\uparrow$ 0\%})\\ 
                & &F1-score & 88.57\%&88.57\% (\textcolor{blue}{$\uparrow$ 0\%})\\ 
                & &$\kappa$ & 83.33\%&83.33\% (\textcolor{blue}{$\uparrow$ 0\%})\\ 
            \cmidrule(lr){2-5}
                &\multirow{3}{*}{Camera 3}&Accuracy& 59.26\%&88.89\% (\textcolor{blue}{$\uparrow$ 29.63\%})\\ 
               & &F1-score & 52.17\%&88.57\% (\textcolor{blue}{$\uparrow$ 36.4\%})\\ 
               & &$\kappa$ & 38.89\%&83.33\% (\textcolor{blue}{$\uparrow$ 44.44\%})\\ 
            \midrule
            \multirow{6}{*}{\textbf{Kiwi}}  &\multirow{3}{*}{Camera 1} &Accuracy&63.89\%&70.83\% (\textcolor{blue}{$\uparrow$ 6.94\%})\\ 
               & &F1-score & 64.55\%&71.11\% (\textcolor{blue}{$\uparrow$ 6.56\%})\\ 
                & &$\kappa$ & 45.83\%&56.25\% (\textcolor{blue}{$\uparrow$ 10.42\%})\\ 
            \cmidrule(lr){2-5}
            &\multirow{3}{*}{Camera 3}&Accuracy& 51.85\%&88.89\% (\textcolor{blue}{$\uparrow$ 37.04\%})\\ 
            & &F1-score & 47.96\%&88.57\% (\textcolor{blue}{$\uparrow$ 40.88\%})\\ 
            & &$\kappa$ & 27.78\%&83.33\% (\textcolor{blue}{$\uparrow$ 55.55\%})\\ 
            \bottomrule
        \end{tabular}
        }
    \end{table}

     \begin{table}
        \centering
        \caption{Detailed performance evaluation for avocado and kiwi (firmness categorie).}
        \label{tab:avocado-kiwi-firmness}
        \resizebox{\columnwidth}{!}{
        \begin{tabular}{lllr>{\columncolor{LightGray}}r}
            \toprule
             &&& DeepHS-Net&\textbf{Fruit-HSNet (Ours)} \\ 
             \midrule
             \multirow{9}{*}{\textbf{Avocado}}  &\multirow{3}{*}{Camera 1} &Accuracy&75.00\%&83.33\% (\textcolor{blue}{$\uparrow$ 8.33\%})\\ 
               & &F1-score & 78.52\%&83.33\% (\textcolor{blue}{$\uparrow$ 4.81\%})\\ 
                & &$\kappa$ & 62.90\%&70.37\% (\textcolor{blue}{$\uparrow$ 7.47\%})\\ 
            \cmidrule(lr){2-5}
                &\multirow{3}{*}{Camera 2}&Accuracy& 96.30\%&100\% (\textcolor{blue}{$\uparrow$ 3.7\%})\\ 
                & &F1-score & 96.70\%&100\% (\textcolor{blue}{$\uparrow$ 3.3\%})\\ 
                & &$\kappa$ & 94.00\%&100\% (\textcolor{blue}{$\uparrow$ 6\%})\\ 
            \cmidrule(lr){2-5}
                &\multirow{3}{*}{Camera 3}&Accuracy& 62.96\%&66.67\% (\textcolor{blue}{$\uparrow$ 3.71\%})\\ 
               & &F1-score & 58.43\%&61.90\% (\textcolor{blue}{$\uparrow$ 3.47\%})\\ 
               & &$\kappa$ & 42.05\%&47.06\% (\textcolor{blue}{$\uparrow$ 5.01\%})\\ 
            \midrule
            \multirow{6}{*}{\textbf{Kiwi}}  &\multirow{3}{*}{Camera 1} &Accuracy&75.36\%&78.26\% (\textcolor{blue}{$\uparrow$ 2.9\%})\\ 
               & &F1-score & 75.56\%&74.58\% (\textcolor{red}{$\downarrow$ 0.98\%})\\ 
                & &$\kappa$ & 58.35\%&60.88\% (\textcolor{blue}{$\uparrow$ 2.53\%})\\ 
            \cmidrule(lr){2-5}
            &\multirow{3}{*}{Camera 3}&Accuracy& 51.85\%&66.67\% (\textcolor{blue}{$\uparrow$ 14.82\%})\\ 
            & &F1-score & 50.47\%&64.13\% (\textcolor{blue}{$\uparrow$ 13.66\%})\\ 
            & &$\kappa$ & 27.78\%&50.00\% (\textcolor{blue}{$\uparrow$ 22.22\%})\\ 
            \bottomrule
        \end{tabular}
        }
    \end{table}

%% file: figures/plots/overall_performance.tex
    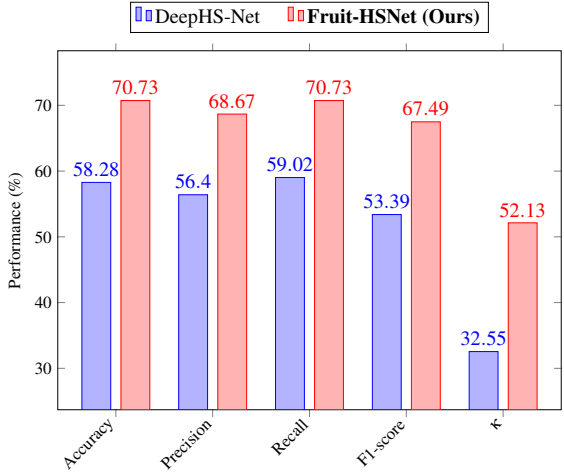
\begin{figure}
    \centering
    \resizebox{\columnwidth}{!}{
    \begin{tikzpicture}
        \begin{axis}[
            width=12cm,height=9cm,
            ybar=6pt, 
            enlargelimits=0.15,
            legend style={
                at={(0.5,1.05)},
                anchor=south,
                legend columns=-1,
                font=\large, 
                /tikz/every even column/.append style={column sep=0.5cm}
            },
            ylabel={Performance (\%)},
            ylabel near ticks,
            symbolic x coords={Accuracy, Precision, Recall, F1-score, $\kappa$},
            xtick=data,
            x tick label style={rotate=45, anchor=east}, 
            bar width=17pt,
            nodes near coords,
            nodes near coords align={vertical},
            every node near coord/.append style={font=\large},
            ymin=30, ymax=72,
            ]
        \addplot coordinates {(Accuracy,58.28) (Precision,56.40) (Recall,59.02) (F1-score,53.39) ($\kappa$,32.55)};
        \addplot coordinates {(Accuracy,70.73) (Precision,68.67) (Recall,70.73) (F1-score,67.49) ($\kappa$,52.13)};
        \legend{DeepHS-Net,\textbf{Fruit-HSNet (Ours)}}
        \end{axis}
    \end{tikzpicture}
    }
    \caption{Performance metrics of Fruit-HSNet across all fruits, hyperspectral cameras, and stages of ripeness.}
    \label{fig:overall-performance}
    \end{figure}

%% file: figures/plots/stratified_performance.tex
\begin{figure*}
    \centering
    \subfloat[By Fruit Type]{%
    \resizebox{0.4\textwidth}{!}{
        \begin{tikzpicture}
            \begin{axis}[
                width=13cm,height=11.1cm,
                ybar=8pt, 
                enlargelimits=0.15,
                legend style={
                    at={(0.35,1.05)},
                    anchor=south,
                    legend columns=-1,
                    font=\Large, 
                    /tikz/every even column/.append style={column sep=0.5cm}
                },
                ylabel={Overall Accuracy (\%)},
                ylabel near ticks,
                symbolic x coords={Avocado, Kiwi, Mango, Kaki, Papaya},
                xtick=data,
                x tick label style={rotate=45, anchor=east,font=\Large}, 
                bar width=17pt,
                nodes near coords,
                nodes near coords align={vertical},
                every node near coord/.append style={font=\Large},
                ymin=40, ymax=82,
                ]
            \addplot coordinates {(Avocado,77.62) (Kiwi,60.11) (Mango,42.59) (Kaki,51.85) (Papaya,62.96)};
            \addplot coordinates {(Avocado,85.19) (Kiwi,71.23) (Mango,65.28) (Kaki,59.72) (Papaya,72.23)};
            \legend{DeepHS-Net,\textbf{Fruit-HSNet (Ours)}}
        \end{axis}
        \end{tikzpicture}
    }
    \label{fig:performance-by-fruit}
    }
    \hfill
    \subfloat[By Camera Type]{%
    \resizebox{0.29\textwidth}{!}{
    \begin{tikzpicture}
        \begin{axis}[
            width=9cm,height=10.3cm,
            ybar=8pt, 
            enlargelimits=0.25,
            ylabel={Overall Accuracy (\%)},
            ylabel near ticks,
            symbolic x coords={Camera 1, Camera 2, Camera 3},
            xtick=data,
            x tick label style={rotate=45, anchor=east,font=\Large}, 
            bar width=17pt,
            nodes near coords,
            nodes near coords align={vertical},
            every node near coord/.append style={font=\Large},
            ymin=50, ymax=70,
            ]
        \addplot coordinates {(Camera 1,58.83) (Camera 2,61.20) (Camera 3,54.81)};
        \addplot coordinates {(Camera 1,69.81) (Camera 2,71.72) (Camera 3,71.11)};
        \end{axis}
    \end{tikzpicture}
    }
    \label{fig:performance-by-camera}
    }\hfill
    \subfloat[By Ripeness Category]{%
    \resizebox{0.29\textwidth}{!}{
    \begin{tikzpicture}
        \begin{axis}[
            width=9cm,height=10.3cm,
            ybar=8pt, 
            enlargelimits=0.25,
            ylabel={Overall Accuracy (\%)},
            ylabel near ticks,
            symbolic x coords={Ripeness, Firmness, Sweetness},
            xtick=data,
            x tick label style={rotate=45, anchor=east,font=\Large}, 
            bar width=17pt,
            nodes near coords,
            nodes near coords align={vertical},
            every node near coord/.append style={font=\Large},
            ymin=50, ymax=72,
            ]
        \addplot coordinates {(Ripeness,56.90) (Firmness,65.10) (Sweetness,53.60) };
        \addplot coordinates {(Ripeness,71.34) (Firmness,75.04) (Sweetness,63.95)};
        \end{axis}
    \end{tikzpicture}
    }
    }
    \caption{Performance Evaluation of Fruit-HSNet for hyperspectral classification of fruit ripeness}
\end{figure*}
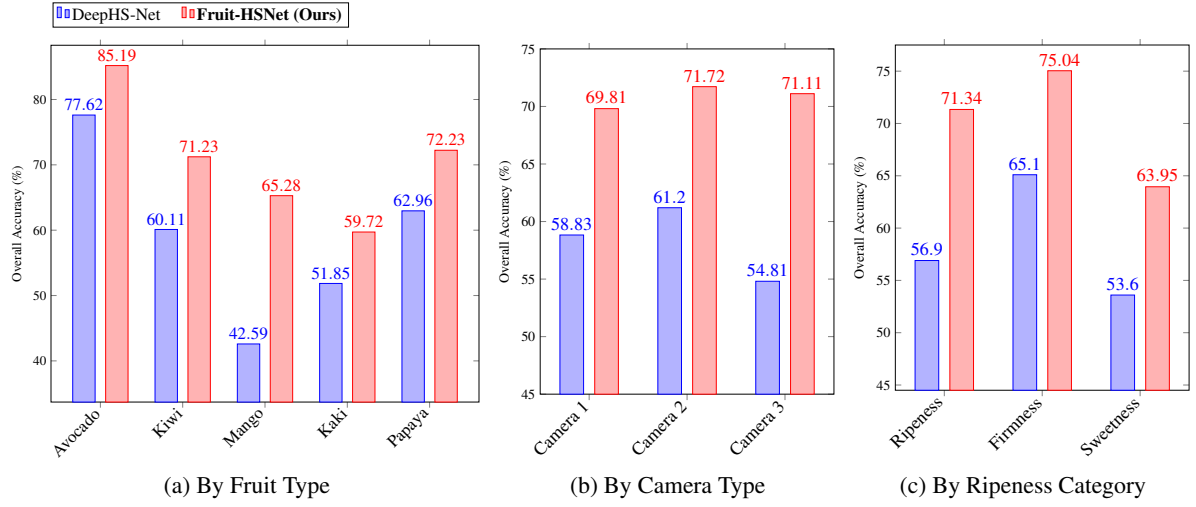 

%% file: sections/04_ablation_study.tex
\subsection{Ablation Study}
In this section, we investigate the influence of key architectural components of Fruit-HSNet on the model's overall performance.
        \begin{table*}
            \centering
            \caption{Feature choices impact on Fruit-HSNet performance. \textit{$^\ast$calculated across all spectral bands.}}
            \label{tab:ablation-feature-extraction}
            \resizebox{\textwidth}{!}{
            \begin{tabular}{lrr}
                \toprule
                 \textbf{Model Variant}&\textbf{Accuracy}&\textbf{Drop in Performance} \\ 
                 \midrule
                  \rowcolor{LightGray} \textbf{Fruit-HSNet (Ours)} &\textbf{70.73} \%&—\\
                 Fruit-HSNet with Wavelet Transform& 52.74 \% &\textcolor{red}{$\downarrow$ 17.99\%}\\  
                 Fruit-HSNet with average of all pixels$^\ast$ & 62.98 \% &\textcolor{red}{$\downarrow$  7.75\%}\\  
                 Fruit-HSNet with variance of all pixels$^\ast$\hspace*{6cm}& 59.92 \% &\textcolor{red}{$\downarrow$  10.81\%}\\  
                \bottomrule
            \end{tabular}
            }
        \end{table*}
        
        \begin{table*}
            \centering
            \caption{Impact of introducing feature fusion module and learnable mechanism on Fruit-HSNet performance. \textit{$^\ast$FE denotes Feature Extraction}.}
            \label{tab:ablation-feature-fusion}
            \resizebox{\textwidth}{!}{
            \begin{tabular}{lrr}
                \toprule
                 \textbf{Model Variant}&\textbf{Accuracy}&\textbf{Drop in Performance} \\ 
                 \midrule
                  \rowcolor{LightGray} \textbf{Fruit-HSNet (Ours)} &\textbf{70.73} \%&— \\
                 Fruit-HSNet w/o spectral FE$^\ast$ module& 48.60 \% &\textcolor{red}{$\downarrow$ 22.13 \%}\\  
                 Fruit-HSNet w/o spatial FE$^\ast$ module&  58.40 \% &\textcolor{red}{$\downarrow$  12.33 \%}\\ 
                 Fruit-HSNet w/o learnable feature fusion\hspace*{6cm}& 60.96 \% &\textcolor{red}{$\downarrow$ 9.77 \%}\\  
                \bottomrule
            \end{tabular}
            }
        \end{table*}
    
    \paragraphheading{Spectral Feature Extraction Module Ablation.}
    To determine the best discriminative spectral feature extraction module for FRP-based hyperspectral image, we compared the performance of Fruit-HSNet using Fourier Transform versus Wavelet Transform in table~\ref{tab:ablation-feature-extraction}.
        
    \paragraphheading{Spatial Feature Extraction Module Ablation.}
    We studied three different methods for extracting spatial features: \textbf{(1)}~central pixel across all spectral bands~(the spectral signature of the central pixel), \textbf{(2)}~mean of all pixels across all spectral bands (the average spectral signature), \textbf{(3)}~variance of all pixels across all spectral bands (the variance in spectral signatures across all pixels) (See Table~\ref{tab:ablation-feature-extraction}).

    \paragraphheading{Feature Fusion Ablation.}
    In this part, we evaluate the performance of concatenating spatial and spectral features compared to using either spatial or spectral features alone. This aims to demonstrate the added value of integrating spectro-spatial features in enhancing the classification accuracy of Fruit-HSNet~(See Table~\ref{tab:ablation-feature-fusion}).

    \paragraphheading{Ablation of Learnable Mechanisms in Feature Fusion.}
    In this part, we compare the efficiency of concatenating learned features versus a simple concatenation (without learning) of features. The learned fusion aims to intelligently combine features in a way that maximizes the relevant information from each feature extraction block~(See Table~\ref{tab:ablation-feature-fusion}).

%% file: sections/05_discussion.tex
\section{\uppercase{Findings and Analysis}}
\paragraphheading{How effective is Fruit-HSNet in hyperspectral image-based fruit ripeness prediction, and how does it compare to state-of-the-art methods?}
The results presented in Table~\ref{tab:overall-performance} and Figure~\ref{fig:overall-performance} demonstrate the effectiveness of Fruit-HSNet for hyperspectral image-based fruit ripeness prediction. Fruit-HSNet produces very promising results despite being trained on a small dataset. The success of our method is due to two factors: the use of \textbf{(1)}~informative and discriminative spatial and spectral features for fruit ripeness classification, and \textbf{(2)}~a learnable feature fusion mechanism that naturally applies attention to the spatio-spectral features, effectively capturing the most relevant features for the candidate fruit, camera, and/or ripeness stage. We detail and discuss our evaluations below.

Fruit-HSNet clearly outperforms other methods, ranging from baseline to state-of-the-art, with an accuracy of~70.73\%. Previous models, which are considered state-of-the-art for this dataset, such as DeepHS-Net, achieve accuracy and F1-score performances of~58.28\% and~53.39\%, respectively, indicating a significant improvement by Fruit-HSNet: an increase of~12.45\% in accuracy and~14.1\% in F1-score. This improvement can be considered promising since the results on this dataset have saturated around the~50\% range, as presented in~\ref{tab:overall-performance}.

Attention/transformer-based methods generally show inferior performance compared to convolution-based methods, suggesting that attention-based architectures do not capture the spatial and spectral features of hyperspectral images as effectively for this specific application. This reaffirms the observation that state-of-the-art methods for hyperspectral image classification, which have shown impressive results in certain applications, fail to maintain comparable performance for different applications, emphasizing the need to adapt attention mechanisms for the specificity of this application.

Analyzing the performance based on the Kappa metric further confirms the superior performance of Fruit-HSNet. While DeepHS-Net shows reasonable accuracy, its Kappa metric is relatively low at~32.55\%. Fruit-HSNet, on the other hand, achieves a Kappa of~52.13\%, representing an improvement of~19.58\%, thereby enhancing the reliability and consistency of predictions in scenarios with imbalanced class distributions, demonstrating the robustness of Fruit-HSNet to variations in input data.

\paragraphheading{What is the performance of Fruit-HSNet specific to each type of fruit?}
So far, we have assessed the performance across all fruits, hyperspectral cameras, and stages of ripeness. We are now analyzing the performance of Fruit-HSNet by fruit type. Table~\ref{tab:fruit-specific-performance} and Figure~\ref{fig:performance-by-fruit} show that Fruit-HSNet consistently outperforms DeepHS-Net across all fruit types (Avocado, Kiwi, Mango, Kaki, Papaya) in terms of all metrics including Overall Accuracy, Average Accuracy, F1-score, and Cohen's kappa coefficient. The improvement margins in Fruit-HSNet over DeepHS-Net are significant.

According to the F1-score, which combines precision and recall, Fruit-HSNet shows an improvement in predicting avocado ripeness: an increase of~8.04\%. For the Kiwi fruit, Fruit-HSNet demonstrates improvements across all metrics as well, with a significant increase in the kappa metric by 16.37\%, indicating a more reliable model performance. For Mango, the least predictable fruit in terms of performance for DeepHS-Net, with an overall accuracy of 42.59\% and a kappa of 3.51\%, we note spectacular improvements with Fruit-HSNet, showing an improvement of 22.69\% and 41.35\% respectively, indicating increased sensitivity of Fruit-HSNet to the spectral characteristics of this particular fruit, and validating the hypothesis of the generalizability of our method across various fruit types. Both Kaki and Papaya show significant improvements, particularly in the F1 and kappa scores, indicating a better balance between precision and recall.

In now analyzing the performance of each fruit by stage of ripeness, Table~\ref{tab:fruit-specific-details} shows that Fruit-HSNet is not only globally more accurate but also better at capturing specific quality attributes of each fruit: For ripeness, we note significant improvements for all fruits, with substantial enhancements for Kiwi and Mango. This indicate a better feature extraction capability of Fruit-HSNet to discern spectral signatures related to ripeness. Regarding firmness, all fruits show performance improvement. In particular, the performance of Kiwi and Mango has significantly increased, due to the increased sensitivity of Fruit-HSNet to repetitive textural attributes detectable by the spatial feature extraction module based on Fourier transform. Sweetness, a more subtle and complex attribute to capture spectrally, also shows improvements, especially in Mango and Kaki.

\paragraphheading{How does the performance of Fruit-HSNet vary with different hyperspectral cameras?}
The performance of Fruit-HSNet was evaluated using three different hyperspectral cameras, each with unique spectral sensitivities. The results, detailed in Tables~\ref{tab:camera-specific-performance} and~\ref{tab:camera-specific-details}, reveal significant variations in model efficiency depending on the camera used, highlighting the crucial impact of imaging hardware on the task of hyperspectral image-based fruit ripeness prediction.

Camera 1 (Specim FX 10) operates in the visible to near-infrared (VNIR) range. With this camera, Fruit-HSNet achieved an overall accuracy of 69.81\%, an improvement of 10.98\% over DeepHS-Net. The average accuracy increased by 11.19\% to 63.98\%, and the F1-score increased by 14\% to 67.12\%. The Kappa coefficient improved by 20.21\% to 51.63\%. These improvements indicate that the spectral information in the VNIR range is effectively used by Fruit-HSNet, enhancing the discrimination of fruit quality attributes such as surface color and certain visible chemical compounds.

Cameras 2 (Corning microHSI 410) and 3 (Innospec Redeye) both operate in the short-wave infrared (SWIR) range, covering wavelengths from 920 to 1730 nm. Despite similar spectral ranges and number of bands, subtle differences in sensor technology and spectral sensitivity may explain the performance variations.

With Camera 2, Fruit-HSNet achieved an overall accuracy of 71.72\%, an improvement of 10.52\%, and an average accuracy of 65.50\%, increased by 10.49\%. The F1-score increased by 12.75\% to 67.67\%, and the Kappa coefficient improved by 16.9\% to 52.09\%. The SWIR range captured by Camera 2 is sensitive to the internal qualities of fruits, such as moisture content and structural properties, which are crucial for assessing attributes like firmness and internal composition.

Camera 3 offered the most significant improvements. Fruit-HSNet achieved an overall accuracy of~71.11\%, representing the highest increase of~16.3\%. The average accuracy improved by~15.75\% to~67.78\%, and the F1-score increased by~17.33\% to~68.15\%. The Kappa coefficient markedly increased by~23.7\% to~53.60\%. Notably, for the ripeness criterion, Camera 3 achieved an exceptional accuracy of~88.89\%, an improvement of~33.33\%. This suggests that the sensor characteristics of Camera 3 are particularly effective at capturing spectral features associated with fruit maturation processes, such as changes in water absorption bands and alterations in chemical composition.

However, performance variations across different quality criteria underscore the influence of camera characteristics on the task of FRP. For firmness, Cameras 1 and 2 achieved higher accuracies (76.76\% and 77.08\%, respectively) compared to 66.67\% for Camera 3. This implies that the spectral features related to fruit properties that influence firmness are better captured by Cameras 1 and 2. Regarding sweetness, Camera 1 obtained the highest accuracy at 70.95\%, showing an improvement of 13.05\%. Camera 2 follows with an accuracy of 61.11\%, while Camera 3 showed a decrease in performance to 44.44\%, indicating a drop of 3.71\%. This decrease cannot be decisive regarding Camera 3, as the sweetness with Camera 3 contains only the kiwi fruit.

This confirms that Fruit-HSNet is generalizable across all types of cameras, as it achieves high results.

\paragraphheading{How robust is Fruit-HSNet for predicting the ripeness of avocados and kiwis?}
In this section, we focus on analyzing the performance of avocados and kiwis given the particular and delicate nature of their ripening process. Tables~\ref{tab:avocado-kiwi-ripeness} and~\ref{tab:avocado-kiwi-firmness} show that Fruit-HSNet accurately predicts the ripeness of avocados and kiwis across different hyperspectral cameras. Compared to DeepHS-Net, Fruit-HSNet demonstrates substantial improvements.

For avocados, Fruit-HSNet showed consistent and superior performance across three cameras:
\textit{(Camera~1)} Fruit-HSNet improved over DeepHS-Net in terms of F1-score, achieving a similarly high accuracy of 83.33\%.
\textit{(Camera~2)} Like DeepHS-Net, Fruit-HSNet demonstrated excellent performance, with an accuracy of 88.89\%, an F1-score of 88.57\%, and a kappa statistic of 83.33\%.
\textit{(Camera~3)} Fruit-HSNet significantly outperformed DeepHS-Net, with a substantial increase of 29.63\% in accuracy, 36.4\% in F1-score, and 44.44\% in kappa statistic.

For kiwis, the performance of Fruit-HSNet was also superior with two cameras:
\textit{(Camera~1)} An increase of 6.94\% in accuracy and 6.56\% in F1-score.
\textit{(Camera~3)} An increase of 37.04\% in accuracy and 40.88\% in F1-score.

\paragraphheading{How does the choice of spectral feature extraction module affect the performance of Fruit-HSNet?}
As shown in Table~\ref{tab:ablation-feature-extraction}, the choice between the Fourier Transform and the Wavelet Transform for spectral feature extraction significantly influences the model's accuracy. The Fourier Transform based feature extraction bloc demonstrated a superior overall accuracy, with an improvement of 17.99\% compared to the Wavelet Transform. This suggests that the periodic patterns of textures and structural changes in the fruit skin extracted by the Fourier Transform are more discernible than those captured by the Wavelet Transform, which focuses on local frequency and time information.

\paragraphheading{How does the choice of spatial feature extraction module influence the performance of Fruit-HSNet?}
Table~\ref{tab:ablation-feature-extraction} presents a comparison of different methods for extracting spatial features. Extracting the spectral signature from the central pixel leads to the highest accuracy, which is an improvement of 7.75\% and 10.81\% over using the mean and variance of pixels per spectral band, respectively. These results underline that the most chemically informative region of the fruit is the center, which is generally indicative of its overall ripeness.

\paragraphheading{What is the significance of employing spatio-spectral features in improving Fruit-HSNet performance?}
The integration of spatial and spectral features is further validated by Table~\ref{tab:ablation-feature-fusion}, where the combination of spatio-spectral features surpasses spatial or spectral features with improvements of 22.13\% and 12.33\%, respectively. This validates the importance of a spatio-spectral approach in the classification of hyperspectral images, as discussed in~\cite{Kumar2024,ahmad2024,frank2023hyperspectral,Muhammad2022}.

\paragraphheading{What impact do learnable mechanisms in feature fusion have on the performance of Fruit-HSNet?} Table~\ref{tab:ablation-feature-fusion} explores the effect of feature fusion with and without learning. Incorporating learning in the feature fusion process led to an improvement of 9.77\%. This improvement emphasizes that the introduction of learnable weights for each feature type allows Fruit-HSNet to adaptively prioritize which type of feature (spectral or spatial) is more informative based on their relevance to fruit ripeness.

%% file: sections/06_conclusion.tex
\section{\uppercase{Conclusion and Future Work}}
In this paper, we introduce Fruit-HSNet, a novel machine learning architecture specifically designed for hyperspectral image-based fruit ripeness prediction. Fruit-HSNet features a small-simple architecture that integrates spatio-spectral feature extraction based on Fourier transform and the central pixel's spectral signature, followed by learnable feature fusion and a deep fully connected neural network.
The experiments conducted on the DeepHS Fruit dataset demonstrated that Fruit-HSNet outperforms existing baselines and state-of-the-art methods across five fruits and three hyperspectral cameras, achieving a new state-of-the-art overall accuracy of 70.73\%.

Future work involves continuous improvement of fruit ripeness prediction results through the integration of an attention mechanism to select the best features. Additionally, for real-world applications, future work will focus on integrating Fruit-HSNet into IoT devices and mobile platforms to facilitate real-time ripeness prediction.

%% file: sections/07_acknowledgements.tex
\section*{\uppercase{Acknowledgements}}
This work was supported by the Agence Universitaire de la Francophonie (AUF) through the IntenSciF program as part of the BIO-Serr (Intelligent Toolbox for Greenhouse Establishment and Monitoring Assistance) Project.